\documentclass[runningheads]{llncs}
\usepackage[T1]{fontenc}
\usepackage{graphicx,verbatim}
\usepackage{multirow}
\usepackage{booktabs} 
\usepackage{amssymb}  
\usepackage{amsmath}
\usepackage{tabularx} 
\usepackage{textcomp}     
\usepackage{marvosym}     
\usepackage{orcidlink}

\usepackage{color}
\usepackage{hyperref}

\begin{document}
\title{GLCP: Global-to-Local Connectivity Preservation for Tubular Structure Segmentation
}
\titlerunning{GLCP: Global-to-Local Connectivity Preservation}
%
\author{Feixiang Zhou 
\inst{1}\orcidlink{0000-0003-4939-9393} \and
Zhuangzhi Gao 
\inst{1}\orcidlink{0009-0000-4339-8088} \and
He Zhao 
\inst{1}\orcidlink{0000-0002-8264-9297}\and
Jianyang Xie 
\inst{1}\orcidlink{0000-0002-4565-5807}\and
Yanda Meng 
\inst{2}\orcidlink{0000-0001-7344-2174}\and
Yitian Zhao 
\inst{3} \and
Gregory Y.H. Lip 
\inst{4} \and
Yalin Zheng 
\inst{1}\orcidlink{0000-0002-7873-0922}\textsuperscript{(\Letter)}
}
\authorrunning{Zhou et al.}
%
\institute{Department of Eye and Vision Sciences, University of Liverpool, Liverpool, UK 
\email{yalin.zheng@liverpool.ac.uk}\\
\and
Department of Computer Science, University of Exeter, Exeter, UK
 \and
Ningbo Institute of Materials Technology and Engineering, CAS, Ningbo, China
\and 
Department of Cardiovascular and Metabolic Medicine, University of Liverpool, Liverpool, UK 
\\
}
\maketitle              
\begin{abstract}
Accurate segmentation of tubular structures, such as vascular networks, plays a critical role in various medical domains. A remaining significant challenge in this task is structural fragmentation, which can adversely impact downstream applications. Existing methods primarily focus on designing various loss functions to constrain global topological structures. However, they often overlook local discontinuity regions, leading to suboptimal segmentation results.
To overcome this limitation, we propose a novel Global-to-Local Connectivity Preservation (GLCP) framework that can simultaneously perceive global and local structural characteristics of tubular networks. Specifically, we propose an Interactive Multi-head Segmentation (IMS) module to jointly learn global segmentation, skeleton maps, and local discontinuity maps, respectively. This enables our model to explicitly target local discontinuity regions while maintaining global topological integrity. In addition, we design a lightweight Dual-Attention-based Refinement (DAR) module to further improve segmentation quality by refining the resulting segmentation maps. Extensive experiments on both 2D and 3D datasets demonstrate that our GLCP achieves superior accuracy and continuity in tubular structure segmentation compared to several state-of-the-art approaches. The source codes will be available at \url{https://github.com/FeixiangZhou/GLCP}.
\keywords{Tubular structure segmentation  \and Vascular segmentation \and Connectivity preservation.}

\end{abstract}
\section{Introduction}

The precise segmentation of thin tubular structures, such as blood vessels, is a fundamental step for many downstream tasks, including disease diagnosis \cite{mou2019cs}, surgical planning \cite{alirr2021automated} and computational biology \cite{ii2020multiscale}. However, segmenting tubular structures is challenging due to their elongated, thin shapes, branching patterns, as well as imaging imperfections such as poor contrast. 

Existing deep learning segmentation methods can be grouped as model-based, feature-based, and loss-based segmentation approaches. Model-based methods \cite{dong2022deu,ma2020rose,qi2023dynamic,yang2022dcu} focus on tailoring network architectures to align with the unique characteristics of tubular structures. However, due to the inherent sparsity of tubular structures in images, these methods often struggle to accurately capture and delineate them. Feature-based methods \cite{li2022global,huang2025representing,zhang2022progressive,qi2022contrastive} aim to capture the geometric and topological characteristics of tubular structures by incorporating additional feature representations into the model, but their performance and efficiency may be compromised by redundant feature representations. Loss-based methods \cite{shit2021cldice,10.1007/978-3-031-72980-5_13,wang2020deep,shi2024centerline,liu2024enhancing,zhao2020improving} develop different loss functions to strengthen the topological connectivity of tubular structures, such as topological constraints based on persistent homology \cite{clough2020topological,hu2019topology} or skeleton-based formulations \cite{shit2021cldice,shi2024centerline}. Despite these advances, current methods tend to focus on global topological constraints while neglecting the fine-grained, local characteristics of discontinuity-prone areas. In \cite{li2023robust}, a topology violation map identifies topological errors in local regions, but its effectiveness in 3D segmentation is uncertain. Additionally, requiring an extra network to refine predictions increases its overall computational complexity. Recent work \cite{huang2025representing} has designed a multi-task paradigm to enhance global connectivity and boundary consistency. However, the interactions between these tasks are limited, and their attention to discontinuities remains insufficient. Therefore, how to identify and correct structural fragmentation, especially in regions prone to discontinuities, remains an open and challenging problem.

\begin{figure*}[tp]
\begin{center}
\includegraphics[width=12cm]{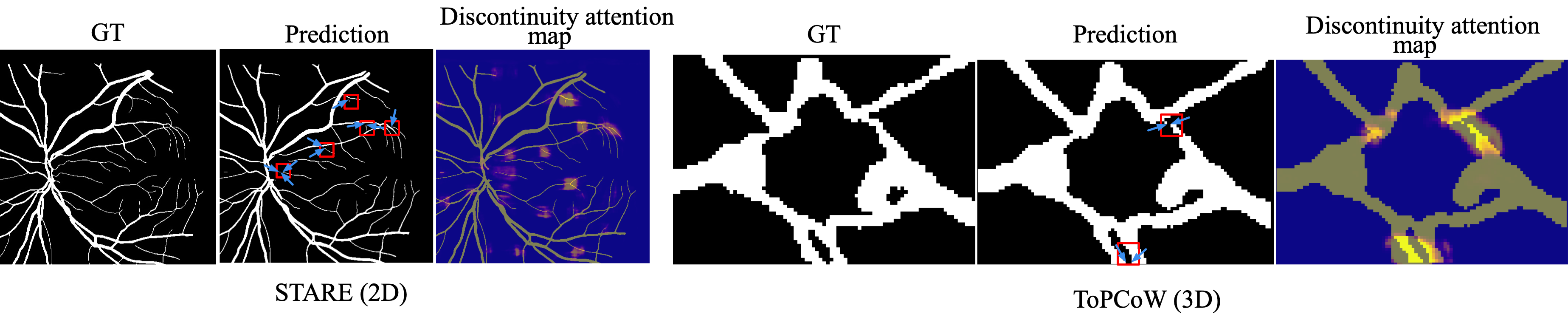}
\end{center}
\caption{Illustration of motivation. In most existing methods, structural fragmentation (red boxes) commonly occurs in predicted masks, manifesting as redundant endpoints (indicated by blue arrows). We propose to directly focus on these regions during training by adaptively learning discontinuity maps, allowing the model to better capture local structural characteristics for connectivity enhancement. For clear visualization, we use the max intensity projection maps of the 3D ground truth (GT) and prediction.}
\label{fig:motivation}
\end{figure*}

\begin{figure*}[tp]
\begin{center}
\includegraphics[width=12cm]{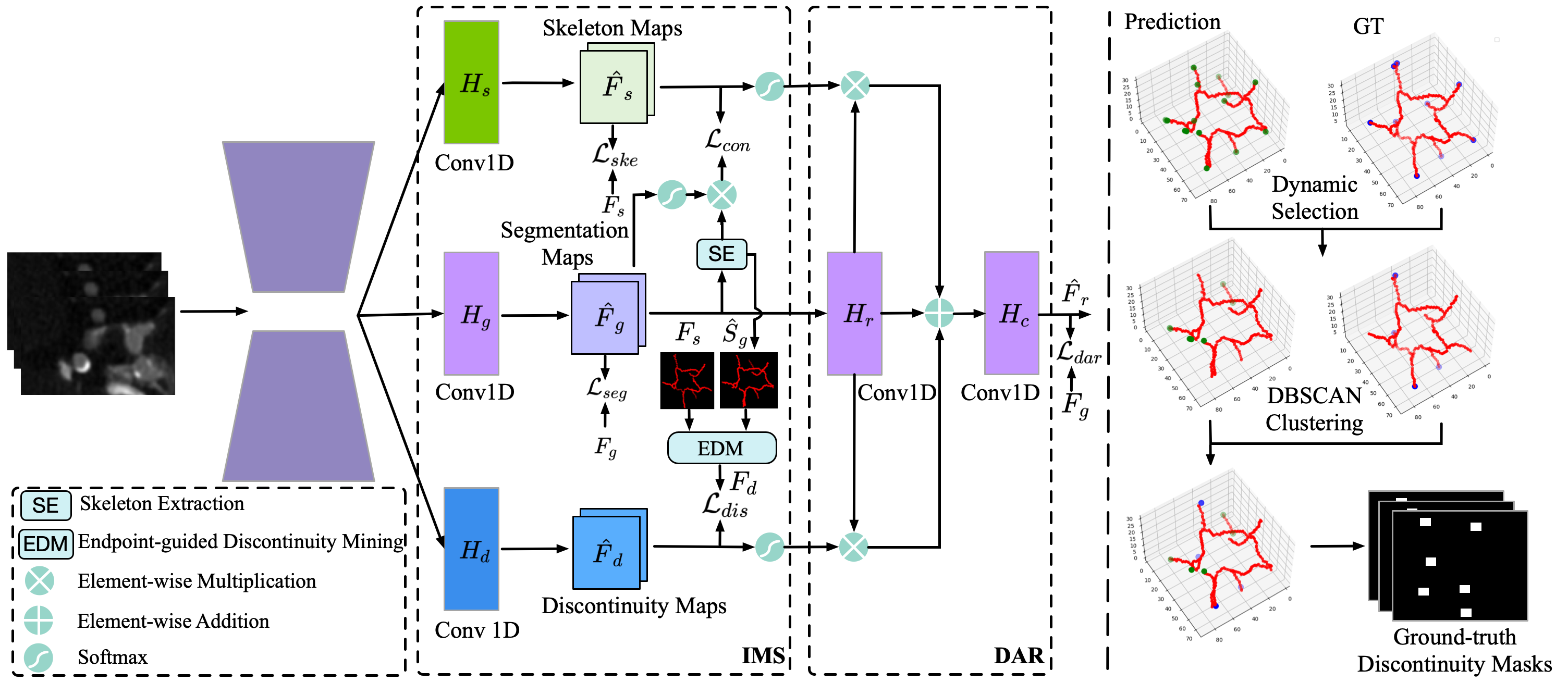}
\end{center}
\caption{An overview of the proposed GLCP. \textbf{Left)} The input patch is first fed into an encoder-decoder backbone to extract image representations. Next, an IMS module uses the encoded representations as input to produce segmentation, discontinuity and skeleton maps, based on which we design a DAR module to integrate global skeleton attention with local discontinuity attention for refining segmentation results. \textbf{Right)} In EDM, endpoints indicating potential discontinuities are identified and used to form the ground-truth discontinuity masks. Note that SE and EDM are only used for training.}
\label{fig:framework}
\end{figure*}

In this paper, we proposed a GLCP framework tailored for both 2D and 3D tasks. Instead of applying topological constraints to the predictions, we proposed Interactive Multi-head Segmentation (IMS) that jointly learns global segmentation, skeleton maps, and local discontinuity maps by a multi-head end-to-end architecture. Specifically, in addition to the segmentation task, we first designed a novel discontinuity prediction task, where a discontinuity head is introduced to predict discontinuity-prone local regions. Intuitively, the structural fragmentation in vascular segmentation results often manifests as redundant endpoints, as shown in Fig. \ref{fig:motivation}. This motivates us to detect these endpoints from the initial predictions (i.e., segmentation maps) to identify potential discontinuity regions, as shown in Fig. \ref{fig:framework},  thereby guiding the model to better focus on local structures by adaptively learning these discontinuities (i.e., discontinuity maps). We also constructed a skeleton prediction task by adding a skeleton head to learn global skeletal representation (i.e., skeleton maps) of foreground objects. In particular, we propose a self-supervised consistency loss to boost the interactions between the main task and skeleton prediction task. Equipped with this multi-head architecture and a shared backbone, our method can better capture global and local structural characteristics of tubular structures, thus improving both segmentation accuracy and topological continuity.

Apart from the IMS, we also introduced a lightweight Dual-Attention-based Refinement (DAR) module to refine the segmentation results based on the initial segmentation, discontinuity and skeleton maps.  More specifically, the probability maps derived from the skeleton and discontinuity maps are utilized as global and local attention maps, guiding the model to refine critical regions, thereby improving the overall segmentation quality.

Our contributions are as follows: \textbf{1)} We propose an IMS strategy to enhance the model’s capability to simultaneously perceive global and local structural characteristics of tubular networks, leading to better connectivity. \textbf{2)} We design a DAR module to refine segmentation results by integrating global and local attention mechanisms based on skeleton and discontinuity maps. \textbf{3)} Extensive experiments on both 2D and 3D datasets demonstrate that our approach outperforms existing methods in terms of both segmentation accuracy and topological continuity.

\section{Methodology}
\label{sec:Method}
An overview of the proposed method is illustrated in Fig. \ref{fig:framework}. Three interactive heads (i.e., $H_g$, $H_d$  and $H_s$) are designed to jointly learn segmentation, discontinuity and skeleton maps (i.e., $\hat{F}_g$, $\hat{F}_d$  and $\hat{F}_s$), allowing our model to pay attention to both global and local structures (Sec. \ref{sec:3.1}). Besides, discontinuity and skeleton attention maps extracted from the corresponding predictions are utilized to further refine the segmentation results (Sec. \ref{sec:3.2}).

\subsection{Interactive Multi-head Segmentation (IMS)}
\label{sec:3.1}

As aforementioned, existing methods lack the ability to perceive local discontinuity regions. To address this limitation, we propose a novel auxiliary task to predict potential discontinuity regions, which enhances the model's awareness of the discontinuities and the robustness of segmentation results. 


Specifically, a new discontinuity head $H_d$ is integrated into the backbone to produce the discontinuity maps $\hat{F}_d$.  However, to enable adaptive learning, the corresponding GT should be carefully constructed. Therefore, we propose an endpoint-guided discontinuity mining strategy to generate ground-truth discontinuity masks. In more detail, we first extract skeletons ($F_s$ and $\hat{S}_g$) from the corresponding ground truth $F_g$ and predicted segmentation masks $\hat{F}_g$, respectively by skeleton extraction algorithms \cite{shit2021cldice,van2014scikit}. 
Given $\hat{S}_g$ and $F_s$, we aim to identify potential structural fragmentation by detecting and analyzing their endpoints. In this way, we first detect all endpoints in $\hat{S}_g$ and $F_s$ through a convolution operation with a fixed kernel, resulting in the sets $\hat{P}_g=\left \{ \hat{p}_1,\hat{p}_2,...,\hat{p}_M \right \} $ and $P_g=\left \{ p_1,p_2,...,p_N \right \}$, respectively, where $M$ and $N$ are the number of endpoints. To identify potential discontinuity regions, we then propose a dynamic endpoint selection strategy, which calculates distances between the endpoints in $\hat{P}_g$ and $P_{g}$, and adaptively selects discontinuity points based on statistical thresholds. Formally, for each $\hat{p}_i  \in \hat{P}_g$, we calculate the shortest Euclidean distance to all the points in $P_{g}$:
\begin{equation}
\begin{split}
\hat{d}_{i} = \min_{p_{j} \in P_{g}} \| \hat{p}_{i} - p_{j} \|_{2},
\label{equation:distance_pred}
\end{split}
\end{equation} 
where $\hat{d}_{i}$ represents the shortest distance from $\hat{p}_{i}$ to $P_{g}$. Then the distances for all endpoints in $\hat{P}_g$ form the set $\hat{D}_{g} = \{\hat{d}_{1}, \hat{d}_{2},..., \hat{d}_{M}\}$.

To adaptively determine discontinuity points, we define a dynamic threshold $\hat{\tau}_{}$ based on the mean and standard deviation of $\hat{D}_{g}$. Any point $\hat{p}_{i} \in \hat{P}_{g}$ with $\hat{d}_{i} > \hat{\tau}$ is then identified as a potential discontinuity point, forming a subset:
\begin{equation}
\begin{split}
\widetilde{P}_{g} = \{\hat{p}_{i} \in \hat{P}_{g} \mid \hat{d}_{i} > \hat{\tau}\}, \hat{\tau} = \operatorname{mean}(\hat{D}_{g}) + \operatorname{std}(\hat{D}_{g})
\label{equation:points_selected_pred}
\end{split}
\end{equation} 
where $\operatorname{mean}(\cdot)$ and $\operatorname{std}(\cdot)$ represent the mean and standard deviation, respectively. Typically, the selected $\widetilde{P}{g}$ includes endpoints that do not exist in the GT but are introduced in the prediction due to discontinuities. However, endpoints present in the GT may also be lost in the prediction. Considering these regions, we also select critical positions from the GT to supplement $\widetilde{P}{g}$. Similarly, we can obtain the set $\overline{P}_{g}$ of discontinuity points from $P_{g}$ using Eqs. (\ref{equation:distance_pred}) and (\ref{equation:points_selected_pred}).
The two sets of discontinuity points are then merged into a unified set $P^{'}  = \widetilde{P}_{g} \cup \overline{P}_{g}$. 
To further refine the selection of discontinuity points, we apply DBSCAN clustering \cite{khan2014dbscan} to $P^{'}$ to obtain a new set $P$, which groups nearby points within a distance, effectively reducing redundancy caused by closely spaced discontinuity points. We randomly select one point from these neighboring points during the training to balance the training data and prevent over-representation of densely clustered discontinuity regions.

Directly predicting sparse and isolated discontinuity points is challenging due to their irregular distribution. Instead, we construct ground-truth discontinuity masks by expanding the identified discontinuity points to their local neighbourhoods. For each discontinuity point $p_{i} \in P$, we define a cube window $R_{i}$ of size $w_z \times w_y \times w_x$, centered on $p_{i}$, which is expressed as:
\begin{equation}
\begin{split}
R_{i} = \{(x, y, z) \mid |x - p_{i}^{x}| \leq w_x/2, |y - p_{i}^{y}| \leq w_y/2,|y - p_{i}^{z}| \leq w_z/2\}
\label{equation:regions}
\end{split}
\end{equation} 
where $p_{i}^{x}$, $p_{i}^{y}$ and $p_{i}^{z}$ are the coordinates of $p_{i}$. Each pixel within this region is assigned with a value of 1, indicating the discontinuity area. The final ground-truth discontinuity mask $F_{f}$ is obtained by taking the union of all such regions, denoted as $F_{d} = \bigcup_{p_{i} \in P} R_{i}$.

Inspired by \cite{huang2025representing}, we also incorporate a skeleton prediction task to complement the discontinuity prediction task. Unlike \cite{huang2025representing} using an additional decoder, we design a skeleton head $H_s$ to predict skeleton maps $\hat{F}_s$, with the corresponding GT denoted as $F_s$. Additionally, we impose self-supervised consistency constraints between global skeletons from this task and the main objective, respectively, to promote their interaction, thus allowing for more accurate and consistent predictions.  The consistency loss $\mathcal{L}_{con}$ can be defined as:
\begin{equation}
\begin{split}
\mathcal{L}_{con}=\operatorname{KL}\big(\sigma(\hat{F}_{g})\otimes \hat{S}_g,\psi(\hat{F}_s)\big)+\operatorname{KL}\big(\hat{F}_s,\psi(\sigma(\hat{F}_{g})\otimes \hat{S}_g)\big)
\label{equation:consistency}
\end{split}
\end{equation} 
where $\sigma(\cdot)$ represents softmax activation function, $\psi(\cdot)$ denotes a function that truncates the gradient of input, and $KL(\cdot)$ is the Kullback-Leibler divergence loss. The proposed consistency loss encourages the two tasks to focus on the same global topological structure by aligning the probability distributions of the skeletons, which promotes stronger inter-task interaction. The use of the gradient truncation function $\psi(\cdot)$ stabilizes the training process by preventing direct interference between the two tasks while still maintaining consistency.


\subsection{Dual-Attention-based Refinement (DAR)}
\label{sec:3.2}

The proposed IMS encourage the model to better capture global and local structural properties by jointly learning segmentation (e.g., $\hat{F}_g$), skeleton (i.e., $\hat{F}_s$) and discontinuity maps (e.g., $\hat{F}_d$). To further improve the segmentation quality, a DAR module is proposed to refine the initial prediction results, e.g., $\hat{F}_g$. Considering the global connectivity and local fragmentation indicated by the skeleton and discontinuity maps, we explicitly construct global and local attention maps to guide the refinement process. Hence, the refined segmentation maps $\hat{F}_{r}$ are formulated as:
\begin{equation} 
\begin{split}
 \hat{F}_{r}=H_c\big (\sigma(\hat{F}_{s})\otimes H_r(\hat{F}_{g})+\sigma(\hat{F}_{d})\otimes H_r(\hat{F}_{g})+H_r(\hat{F}_{g})\big)
\label{equation:refine}
\end{split}
\end{equation} 
where $\sigma(\cdot)$ is used to yield skeleton and discontinuity attention maps. $H_{r}(\cdot)$ and $H_{c}(\cdot)$ denote convolution operations with the kernel size of 1, where $H_{r}(\cdot)$ maps $\hat{F}_{g}$ to a higher-dimensional space, and $H_{c}(\cdot)$ generates the final prediction results.

By integrating the DAR with IMS, the overall loss function is calculated as:
\begin{equation} 
\begin{split}
\mathcal{L}_{total}=\mathcal{L}_{ims}  +\alpha \mathcal{L}_{con}+\beta \mathcal{L}_{dar}  
\label{equation:total_loss}
\end{split}
\end{equation}
where $\alpha$ and $\beta$ are weight factors. $\mathcal{L}_{ims}$ represents the sum of the losses of the segmentation ($\mathcal{L}_{seg}$), discontinuity($\mathcal{L}_{dis}$) and skeleton ($\mathcal{L}_{ske}$) predictions. $\mathcal{L}_{dar}$ denotes the refinement loss. We use cross-entropy (CE) loss for $\mathcal{L}_{dar}$, while $\mathcal{L}_{seg}$, $\mathcal{L}_{dis}$ and $\mathcal{L}_{ske}$ follow the default loss functions in nnUNet \cite{isensee2021nnu}.

\section{Experiments}
\subsection{Datasets and Evaluation Metrics}
We evaluate our proposed method on three benchmark datasets. The STARE dataset \cite{hoover2003locating}  is used for 2D retinal vessel segmentation, containing 10 images for training and 10 for testing. The CCA dataset \cite{yang2024segmentation} 
provides 20 CTA images depicting coronary artery disease, with 16 for training and 4 for testing. The MICCAI 2023 TopCoW Challenge dataset \cite{yang2024benchmarking} consists of 90 brain CTA cases, with 72 for training and 18 for testing. Both binary and multi-class settings on TopCoW are employed in this work. We report the  volumetric scores (Dice and clDice \cite{shit2021cldice}), topology errors (Betti Error \cite{hu2019topology} $\beta$ for the sum of Betti Numbers $\beta_{0}$ and $\beta_{1}$) and distance errors (Hausdorff Distance (HD) \cite{taha2015metrics}).

\subsection{Implementation Details}
Similar to \cite{shi2024centerline}, nnUNet V2 is selected as our baseline. The proposed GLCP can be seamlessly integrated with the baseline by adding only four lightweight convolution layers ($H_{d}$, $H_{s}$, $H_{r}$ and $H_{c}$) at the end, without introducing significant modifications. $R_i$ is set to one-eighth of the input patch size, and both $\alpha$ and $\beta$ are set to 0.5. For other training and optimization settings, we follow the default configurations of nnUNet for each dataset. In the skeleton prediction task of the multi-class setting, all classes except the background are treated as the foreground object to extract the overall skeleton.

\begin{table*}[ht]
\centering
\caption{Comparison with the state-of-the-art methods on the STARE, ToPCoW-binary and CCA datasets. * means training with default CE and Dice loss. MD means multi-decoder configuration \cite{huang2025representing} consisting of skeleton and edge prediction tasks.}
\label{tab:comparison}
\resizebox{1\textwidth}{!}{
\begin{tabular}
{c|cccc|cccc|cccc}
\toprule
\multirow{2}{*}{Method} & \multicolumn{4}{c|}{STARE}& \multicolumn{4}{c|}{ToPCoW-binary}& \multicolumn{4}{c}{CCA}   
\\ \cmidrule{2-13}
& \textbf{Dice}$\uparrow $ & \textbf{clDice}$\uparrow $ & $\beta$$\downarrow $ & \textbf{HD}$\downarrow$  & 

\textbf{Dice}$\uparrow$ & \textbf{clDice}$\uparrow$ & $\beta$$\downarrow $ & \textbf{HD}$\downarrow$ &

\textbf{Dice}$\uparrow$ & \textbf{clDice}$\uparrow$ & $\beta$$\downarrow $ & \textbf{HD}$\downarrow$\\ \midrule
 nnUNet* \cite{isensee2021nnu} & 82.92 &86.25 &4.60 & 6.94 & 90.43 & 95.41 & 1.94 & 1.68 & 86.73 & 87.53 & 35.75 & 21.30  \\
+clDice \cite{shit2021cldice} & 83.30 & 86.99 & 3.90 & 5.30 & 90.63 & 95.57 & 1.72 & 1.63 & 87.57 & 88.19 & 19.50 & 19.27 \\
+cbDice \cite{shi2024centerline} &83.05 & 86.34 & 4.50 & 5.17 & 90.41 & 95.44 & 1.67 & 1.65 & 86.74 & 87.62 & 17.25 & 26.53 \\
+ ske-recall \cite{10.1007/978-3-031-72980-5_13} & 83.39  & 87.11 & 3.70  & 5.06 & 91.02 & 95.44 & 1.56 & \textbf{1.58} & 87.84 & 89.50 & 15.00 & 16.42 \\
+ Ours (GLCP) & \textbf{83.67} & \textbf{87.44} & \textbf{3.00} & \textbf{4.64} & \textbf{91.34} & \textbf{95.58} & \textbf{1.17} & 1.60 & \textbf{87.94}&\textbf{ 90.55} & \textbf{10.50} & \textbf{14.03} \\
\cmidrule{1-13} 
+ MD \cite{huang2025representing}& 83.26 & 87.02 & 4.10 & 5.28  & 90.84 & 95.43 & 1.61 & 1.61  & 87.32 & 88.01 & 25.75 & 16.69 \\
+ Ours (IMS) & \textbf{83.58} & \textbf{87.42} & \textbf{3.20} & \textbf{4.74}  & \textbf{91.24} & \textbf{95.52} & \textbf{1.39} & \textbf{1.53} & \textbf{88.09} & \textbf{89.95} & \textbf{12.25} & \textbf{14.26}\\
\cmidrule{1-13} 
SwinUNETR* \cite{hatamizadeh2021swin}& 82.16 & 86.12 & 5.10 & 7.21  & 90.15 & 95.42 & 2.00 & 1.84  & 85.44 & 86.45 & 44.25 &26.34 \\
+ Ours (GLCP) & \textbf{83.25} & \textbf{86.96} & \textbf{3.60} & \textbf{4.96}  & \textbf{91.27} & \textbf{95.52} & \textbf{1.24} & \textbf{1.64} & \textbf{87.12} & \textbf{89.63} & \textbf{17.50} & \textbf{20.25}\\
\bottomrule
\end{tabular}}
\end{table*}

\subsection{Comparison Results and Ablation Study}

\begin{table*}[ht]
\centering
\begin{minipage}{0.45\linewidth}
\caption{Comparison with the state-of-the-art methods on ToPCoW-multi.}
\label{tab:comparison_multi}
\resizebox{0.95\textwidth}{1.8cm}{
\begin{tabular}{c|cccc}
\toprule
Method & \textbf{Dice}$\uparrow $ & \textbf{clDice}$\uparrow $ & $\beta$$\downarrow $ & \textbf{HD}$\downarrow$ \\ \midrule
 nnUNet* \cite{isensee2021nnu} & 71.15 & 86.84 & 0.57 & 3.21 \\
+clDice \cite{shit2021cldice} & 71.41 & 88.62 & 0.50 & 3.02\\
+cbDice \cite{shi2024centerline} & 71.55 & 88.49 & 0.48 & 3.14 \\
+ ske-recall \cite{10.1007/978-3-031-72980-5_13} & 72.26 & 88.63 & 0.49 & 2.70\\
+ Ours (GLCP)  & \textbf{74.22} & \textbf{89.15} & \textbf{0.39} & \textbf{2.64}\\
\cmidrule{1-5}
+ MD \cite{huang2025representing} & 72.12 & 88.83 & 0.51 & 2.82\\
+ Ours (IMS) & \textbf{73.33} & \textbf{89.03} & \textbf{0.41} & \textbf{2.79}\\
\cmidrule{1-5}
SwinUNETR* \cite{hatamizadeh2021swin} & 70.12 & 86.13 & 0.63 & 3.30\\
+ Ours (GLCP) & \textbf{72.45} & \textbf{88.54} & \textbf{0.44} & \textbf{2.84}\\
\bottomrule
\end{tabular}}
\end{minipage}
\hfill
\begin{minipage}{0.5\linewidth}
 \caption{Ablation study of components.}
    \label{tab:ablation_results}
    \resizebox{1\textwidth}{2.0cm}{
    \begin{tabular}{l|cccc|cccc}
        \toprule
        \textbf{Dataset} & \textbf{Ske} & \textbf{Dis} & \textbf{Self}& \textbf{DAR}& \textbf{Dice}$\uparrow$  & \textbf{clDice}$\uparrow$  & \textbf{$\beta$}$\downarrow$  & \textbf{HD}$\downarrow$ \\
        \midrule
        \multirow{6}{*}{STARE}  &  &  &  & & 82.92 &86.25 &4.60 & 6.94 \\
         & \checkmark & &  &  & 83.16 & 86.31 & 4.40 & 6.22 \\
         &  & \checkmark &  &  & 83.54 & 86.72 & 4.10 & 4.89 \\
         & \checkmark & \checkmark &  &  & 83.45 &87.08 & 3.60 & 5.62 \\
         & \checkmark & \checkmark & \checkmark &  & 83.58 & 87.42 & 3.20 & 4.74 \\
         & \checkmark & \checkmark & \checkmark & \checkmark & \textbf{83.67} & \textbf{87.44} & \textbf{3.00} & \textbf{4.64} \\
        \cmidrule(lr){1-9}
        \multirow{6}{*}{\shortstack{ToPCoW\\-multi}} &  &  &  &  & 71.15 & 86.84 & 0.57 & 3.21 \\
         & \checkmark &  &  & & 71.59 & 88.50 & 0.54 & 3.52  \\
         & & \checkmark &  & & 72.38 & 88.78 & 0.49 & 2.82 \\
         & \checkmark & \checkmark &  & & 73.26 & \textbf{89.57} & 0.43 & 2.85 \\
         & \checkmark & \checkmark & \checkmark & & 73.33 & 89.03 & 0.41 & 2.79 \\
         & \checkmark & \checkmark & \checkmark &  \checkmark & \textbf{74.22} & 89.15 & \textbf{0.39} & \textbf{2.64} \\
        \bottomrule
    \end{tabular}}
\end{minipage}
\end{table*}

Table \ref{tab:comparison} presents the quantitative results of our method compared to five other SOTA loss functions. We use the default loss weights provided in their publicly available codes to implement these methods. For binary segmentation tasks on STARE, ToP-CoW and CCA, our proposed GLCP (IMS+DAR) achieves improvement in terms of Dice and clDice scores, compared to the closest competitors. Meanwhile, from a topological perspective, our GLCP demonstrates superior topological continuity, achieving the lowest $\beta$ errors of 3.00, 1.17, and 10.50, respectively. These results highlight the ability of our method to effectively capture both global and local structural characteristics of 2D and 3D tubular networks, delivering more accurate segmentation performance and ensuring a more continuous topology. In addition, our method achieves competitive results in the quality of boundary extraction, attaining the lowest HD on both the STARE and CCA datasets. We also compare our IMS with a similar multi-task learning framework proposed in \cite{huang2025representing}. For a fair comparison, we re-implement the multi-decoder (MD) prediction network from \cite{huang2025representing} based on the nnUNet framework. The results demonstrate that our IMS significantly outperforms MD across all metrics. In addition to binary segmentation tasks, we also evaluated our method on multi-class segmentation tasks, as shown in Table \ref{tab:comparison_multi}. Our GLCP achieves the best Dice of 74.22\% and clDice of 89.15\% among all the methods in comparison,  representing improvements of 3.07\% and 2.31\% over the baseline, respectively. In terms of topology and distance errors, our method also outperforms these approaches, demonstrating superior topological continuity and boundary accuracy. Finally, to validate the generalizability of our method, we apply GLCP to the transformer-based SwinUNETR \cite{hatamizadeh2021swin}, as shown in the last two rows of Tables \ref{tab:comparison} and \ref{tab:comparison_multi}, demonstrating its effectiveness across different architectures.

\begin{figure*}[tp]
\begin{center}
\includegraphics[width=12cm]{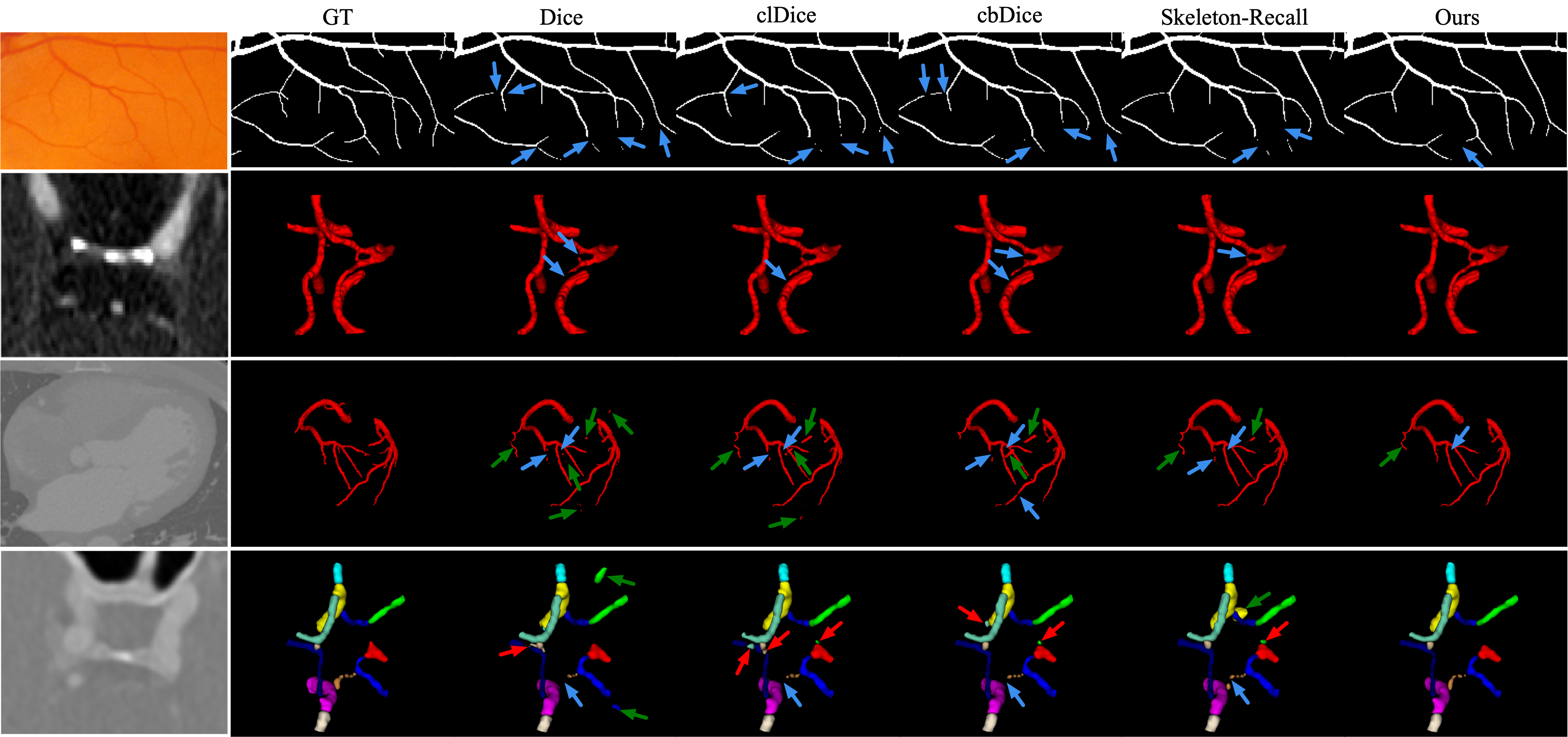}
\end{center}
\caption{Qualitative results on the STARE, ToPCoW-binary, CCA and ToPCoW-multi datasets (listed from top to bottom). Blue, green, and red arrows indicate regions of false negatives, false positives, and misclassifications, respectively, highlighting the challenges in segmentation and the improvements achieved by our method.
}
\label{fig:vis_results}
\end{figure*}

To better illustrate the effectiveness of our method, we showcase several qualitative results on both 2D and 3D examples in Fig. \ref{fig:vis_results}. The results demonstrate that our method is more effective than other approaches in eliminating topology errors and misclassifications when compared to the ground truth.

Table 3 summarizes the ablation study results on the STARE and ToPCoW-multi datasets, highlighting the effectiveness of each proposed component. Specifically, we can observe that the discontinuity prediction task (Dis) plays a more important role in correcting topology errors than the skeleton prediction task (Ske) and combining them can gain further improvement. 
Besides, incorporating the self-supervised (self) consistency constraints loss achieves slight improvements across nearly all the metrics. This can be attributed to the ability of this loss to enhance inter-task interactions by boosting the alignment between the skeletons from the main task and the skeleton prediction task. Finally, the DAR module contributes to further improvements by refining the segmentation maps based on the proposed dual-attention mechanism, achieving the best performance across Dice, $\beta$ error and HD.

\section{Conclusions}
In this paper, we proposed a novel GLCP framework to address the challenge of structural fragmentation in tubular structure segmentation, particularly for vascular networks. We introduced IMS to jointly perceive global topological structures and local discontinuity regions, ensuring improved segmentation accuracy and continuity. Additionally, DAR is proposed to refine segmentation quality through the dual-attention mechanism. Extensive experiments on 2D and 3D datasets demonstrate that our method achieves state-of-the-art performance. Future research will explore the generalizability of our approach across additional segmentation networks and validate its effectiveness on larger and more complex vascular datasets.

\begin{credits}
\subsubsection{\ackname} 
This study was funded by TARGET, a project funded by the 
EU HORIZON EUROPE framework programme for research and innovation under the Grant Agreement No. 101136244.

\subsubsection{\discintname}
The authors have no competing interests to declare that are relevant to the content of this article.
\end{credits}

%
%
%
\bibliographystyle{splncs04}
\bibliography{Paper-1868}

\begin{thebibliography}{10}
\providecommand{\url}[1]{\texttt{#1}}
\providecommand{\urlprefix}{URL }
\providecommand{\doi}[1]{https://doi.org/#1}

\bibitem{alirr2021automated}
Alirr, O.I., Abd~Rahni, A.A.: An automated liver vasculature segmentation from ct scans for hepatic surgical planning. International Journal of Integrated Engineering  \textbf{13}(1),  188--200 (2021)

\bibitem{clough2020topological}
Clough, J.R., Byrne, N., Oksuz, I., Zimmer, V.A., Schnabel, J.A., King, A.P.: A topological loss function for deep-learning based image segmentation using persistent homology. IEEE transactions on pattern analysis and machine intelligence  \textbf{44}(12),  8766--8778 (2020)

\bibitem{dong2022deu}
Dong, S., Pan, Z., Fu, Y., Yang, Q., Gao, Y., Yu, T., Shi, Y., Zhuo, C.: Deu-net 2.0: Enhanced deformable u-net for 3d cardiac cine mri segmentation. Medical Image Analysis  \textbf{78},  102389 (2022)

\bibitem{hatamizadeh2021swin}
Hatamizadeh, A., Nath, V., Tang, Y., Yang, D., Roth, H.R., Xu, D.: Swin unetr: Swin transformers for semantic segmentation of brain tumors in mri images. In: International MICCAI brainlesion workshop. pp. 272--284. Springer (2021)

\bibitem{hoover2003locating}
Hoover, A., Goldbaum, M.: Locating the optic nerve in a retinal image using the fuzzy convergence of the blood vessels. IEEE transactions on medical imaging  \textbf{22}(8),  951--958 (2003)

\bibitem{hu2019topology}
Hu, X., Li, F., Samaras, D., Chen, C.: Topology-preserving deep image segmentation. Advances in neural information processing systems  \textbf{32} (2019)

\bibitem{huang2025representing}
Huang, J., Zhou, Y., Luo, Y., Liu, G., Guo, H., Yang, G.: Representing topological self-similarity using fractal feature maps for accurate segmentation of tubular structures. In: European Conference on Computer Vision. pp. 143--160. Springer (2025)

\bibitem{ii2020multiscale}
Ii, S., Kitade, H., Ishida, S., Imai, Y., Watanabe, Y., Wada, S.: Multiscale modeling of human cerebrovasculature: A hybrid approach using image-based geometry and a mathematical algorithm. PLoS computational biology  \textbf{16}(6),  e1007943 (2020)

\bibitem{isensee2021nnu}
Isensee, F., Jaeger, P.F., Kohl, S.A., Petersen, J., Maier-Hein, K.H.: nnu-net: a self-configuring method for deep learning-based biomedical image segmentation. Nature methods  \textbf{18}(2),  203--211 (2021)

\bibitem{khan2014dbscan}
Khan, K., Rehman, S.U., Aziz, K., Fong, S., Sarasvady, S.: Dbscan: Past, present and future. In: The fifth international conference on the applications of digital information and web technologies (ICADIWT 2014). pp. 232--238. IEEE (2014)

\bibitem{10.1007/978-3-031-72980-5_13}
Kirchhoff, Y., Rokuss, M.R., Roy, S., Kovacs, B., Ulrich, C., Wald, T., Zenk, M., Vollmuth, P., Kleesiek, J., Isensee, F., Maier-Hein, K.: Skeleton recall loss for connectivity conserving and resource efficient segmentation of thin tubular structures. In: Leonardis, A., Ricci, E., Roth, S., Russakovsky, O., Sattler, T., Varol, G. (eds.) Computer Vision -- ECCV 2024. pp. 218--234. Springer Nature Switzerland, Cham (2024)

\bibitem{li2023robust}
Li, L., Ma, Q., Ouyang, C., Li, Z., Meng, Q., Zhang, W., Qiao, M., Kyriakopoulou, V., Hajnal, J.V., Rueckert, D., et~al.: Robust segmentation via topology violation detection and feature synthesis. In: International Conference on Medical Image Computing and Computer-Assisted Intervention. pp. 67--77. Springer (2023)

\bibitem{li2022global}
Li, Y., Zhang, Y., Liu, J.Y., Wang, K., Zhang, K., Zhang, G.S., Liao, X.F., Yang, G.: Global transformer and dual local attention network via deep-shallow hierarchical feature fusion for retinal vessel segmentation. IEEE Transactions on Cybernetics  \textbf{53}(9),  5826--5839 (2022)

\bibitem{liu2024enhancing}
Liu, C., Ma, B., Ban, X., Xie, Y., Wang, H., Xue, W., Ma, J., Xu, K.: Enhancing boundary segmentation for topological accuracy with skeleton-based methods. arXiv preprint arXiv:2404.18539  (2024)

\bibitem{ma2020rose}
Ma, Y., Hao, H., Xie, J., Fu, H., Zhang, J., Yang, J., Wang, Z., Liu, J., Zheng, Y., Zhao, Y.: Rose: a retinal oct-angiography vessel segmentation dataset and new model. IEEE transactions on medical imaging  \textbf{40}(3),  928--939 (2020)

\bibitem{mou2019cs}
Mou, L., Zhao, Y., Chen, L., Cheng, J., Gu, Z., Hao, H., Qi, H., Zheng, Y., Frangi, A., Liu, J.: Cs-net: Channel and spatial attention network for curvilinear structure segmentation. In: Medical Image Computing and Computer Assisted Intervention--MICCAI 2019: 22nd International Conference, Shenzhen, China, October 13--17, 2019, Proceedings, Part I 22. pp. 721--730. Springer (2019)

\bibitem{qi2022contrastive}
Qi, X., Yang, G., He, Y., Liu, W., Islam, A., Li, S.: Contrastive re-localization and history distillation in federated cmr segmentation. In: International Conference on Medical Image Computing and Computer-Assisted Intervention. pp. 256--265. Springer (2022)

\bibitem{qi2023dynamic}
Qi, Y., He, Y., Qi, X., Zhang, Y., Yang, G.: Dynamic snake convolution based on topological geometric constraints for tubular structure segmentation. In: Proceedings of the IEEE/CVF International Conference on Computer Vision. pp. 6070--6079 (2023)

\bibitem{shi2024centerline}
Shi, P., Hu, J., Yang, Y., Gao, Z., Liu, W., Ma, T.: Centerline boundary dice loss for vascular segmentation. In: International Conference on Medical Image Computing and Computer-Assisted Intervention. pp. 46--56. Springer (2024)

\bibitem{shit2021cldice}
Shit, S., Paetzold, J.C., Sekuboyina, A., Ezhov, I., Unger, A., Zhylka, A., Pluim, J.P., Bauer, U., Menze, B.H.: cldice-a novel topology-preserving loss function for tubular structure segmentation. In: Proceedings of the IEEE/CVF conference on computer vision and pattern recognition. pp. 16560--16569 (2021)

\bibitem{taha2015metrics}
Taha, A.A., Hanbury, A.: Metrics for evaluating 3d medical image segmentation: analysis, selection, and tool. BMC medical imaging  \textbf{15},  1--28 (2015)

\bibitem{van2014scikit}
Van~der Walt, S., Sch{\"o}nberger, J.L., Nunez-Iglesias, J., Boulogne, F., Warner, J.D., Yager, N., Gouillart, E., Yu, T.: scikit-image: image processing in python. PeerJ  \textbf{2}, ~e453 (2014)

\bibitem{wang2020deep}
Wang, Y., Wei, X., Liu, F., Chen, J., Zhou, Y., Shen, W., Fishman, E.K., Yuille, A.L.: Deep distance transform for tubular structure segmentation in ct scans. In: Proceedings of the IEEE/CVF Conference on Computer Vision and Pattern Recognition. pp. 3833--3842 (2020)

\bibitem{yang2024benchmarking}
Yang, K., Musio, F., Ma, Y., Juchler, N., Paetzold, J.C., Al-Maskari, R., H{\"o}her, L., Li, H.B., Hamamci, I.E., Sekuboyina, A., et~al.: Benchmarking the cow with the topcow challenge: Topology-aware anatomical segmentation of the circle of willis for cta and mra. ArXiv pp. arXiv--2312 (2024)

\bibitem{yang2024segmentation}
Yang, X., Xu, L., Yu, S., Xia, Q., Li, H., Zhang, S.: Segmentation and vascular vectorization for coronary artery by geometry-based cascaded neural network. IEEE Transactions on Medical Imaging  (2024)

\bibitem{yang2022dcu}
Yang, X., Li, Z., Guo, Y., Zhou, D.: Dcu-net: A deformable convolutional neural network based on cascade u-net for retinal vessel segmentation. Multimedia Tools and Applications  \textbf{81}(11),  15593--15607 (2022)

\bibitem{zhang2022progressive}
Zhang, X., Zhang, J., Ma, L., Xue, P., Hu, Y., Wu, D., Zhan, Y., Feng, J., Shen, D.: Progressive deep segmentation of coronary artery via hierarchical topology learning. In: International Conference on Medical Image Computing and Computer-Assisted Intervention. pp. 391--400. Springer (2022)

\bibitem{zhao2020improving}
Zhao, H., Li, H., Cheng, L.: Improving retinal vessel segmentation with joint local loss by matting. Pattern Recognition  \textbf{98},  107068 (2020)

\end{thebibliography}

\end{document}